\documentclass[10pt,twocolumn,letterpaper]{article}

\usepackage[pagenumbers]{cvpr} 
\makeatletter
\@namedef{ver@everyshi.sty}{}
\makeatother

\usepackage{graphicx}
\usepackage{amsmath}
\usepackage{amssymb}
\usepackage{booktabs}
\usepackage{float}
\usepackage{array}
\newcolumntype{P}[1]{>{\centering\arraybackslash}m{#1}}
\usepackage[table]{xcolor}
\usepackage{enumitem}
\usepackage{subcaption}
\usepackage[pagebackref,breaklinks,colorlinks]{hyperref}

\usepackage[capitalize]{cleveref}
\crefname{section}{Sec.}{Secs.}
\Crefname{section}{Section}{Sections}
\Crefname{table}{Table}{Tables}
\crefname{table}{Tab.}{Tabs.}

\usepackage{xcolor}
\usepackage{tikz}
\usetikzlibrary{quantikz}

\definecolor{my_purple}{HTML}{3F0359}
\definecolor{my_green}{HTML}{A6D95B}
\definecolor{my_yellow}{HTML}{A9D98F} 
\definecolor{my_orange}{HTML}{9579d1} 
\definecolor{my_blue}{HTML}{7eb8da} 
\definecolor{my_teal}{HTML}{4D8C86}

\def\cvprPaperID{9} 
\def\confName{CVPR}
\def\confYear{2023}

\begin{document}

\title{Schrödinger's Camera: First Steps Towards a \\ Quantum-Based Privacy Preserving Camera}

\author{Hannah Kirkland \\
University of Florida\\
{\tt\small hkirkland@ufl.edu}
\and Sanjeev J. Koppal \\
University of Florida\\
{\tt\small sjkoppal@ece.ufl.edu}
}
\maketitle

\begin{abstract}
Privacy-preserving vision must overcome the dual challenge of utility and privacy. Too much anonymity renders the images useless, but too little privacy does not protect sensitive data. We propose a novel design for privacy preservation, where the imagery is stored in quantum states. In the future, this will be enabled by quantum imaging cameras, and, currently, storing very low resolution imagery in quantum states is possible. Quantum state imagery has the advantage of being both private and non-private till the point of measurement. This occurs even when images are manipulated, since every quantum action is fully reversible. We propose a control algorithm, based on double deep Q-learning, to learn how to anonymize the image before measurement. After learning, the RL weights are fixed, and new attack neural networks are trained from scratch to break the system's privacy. Although all our results are in simulation, we demonstrate, with these first steps, that it is possible to control both privacy and utility in a quantum-based manner.  
\end{abstract}

\section{Quantum-Based Privacy for Vision}
\label{sec:intro}

While machine learning algorithms extract useful knowledge from massive amounts of data, there exist concerns that such techniques can infer sensitive, private information without permission. One of the goals of privacy-preserving computer vision is to allow the machine learning algorithm to access visual information useful for a desired task, but not access additional sensitive information~\cite{hoyle2020privacy}.

A major challenge in privacy-preserving computer vision is the privacy-utility trade-off. Blank images are private but regress to data priors for any vision task. Clear images, while useful, do not provide an acceptable level of trust and privacy.  

\begin{figure}[h]
    \centering
        \begin{tikzpicture}
        \node[scale=0.8] at (0,0) {
        \begin{tikzpicture}
            \filldraw[rounded corners, fill=black!10, draw=black, thick] (0,0) rectangle (4,8);
            \node at (2,8) [above] {Environment};
            \filldraw[rounded corners, fill=black!10, draw=black, thick] (5.5,2.75) rectangle (9.5,8);
            \node at (7.5,8) [above] {Agent};
            
            \node[scale=0.4] at (2, 7) {
                \begin{tikzpicture}
                
                    \filldraw[thick, fill=my_blue!95]  (0,0) -- (1,0) -- (1,3) -- (0,3) -- (0,0);
        
                    \filldraw[thick, fill=my_blue!65]  (0.6,3.3) -- (1,3.3) -- (1,3) -- (0.6,3) -- (0.6,3.3);
                    
                    \filldraw[thick, fill=my_blue!95]  (-0.5,1) -- (0,1) -- (0,2) -- (-0.5,2) -- (-0.5,1);
                    
                    \filldraw[thick, fill=my_blue!65]  (1,0) rectangle (6,3) node[font=\LARGE, midway,text width=4.5cm, align=center] {Quantum Camera\\System};
                    
                \end{tikzpicture}
             };
             \draw [-stealth] (2, 6.35) -- (2, 5.8);

             \filldraw[fill=my_yellow] (1.25, 4.3) rectangle (2.75, 5.8) node[midway] {Image};
             
             \draw [-stealth] (2, 4.3) -- (1, 3.8);
             \draw [-stealth] (2, 4.3) -- (3, 3.8);

             \filldraw[rounded corners, fill=my_orange!90] (0.25, 1.8) rectangle (1.75, 3.8) node[below left=0.08in, align=center] {Private\\CNN};

             \filldraw[rounded corners, fill=my_orange!90] (2.25, 1.8) rectangle (3.75, 3.8) node[below left=0.1in, align=center] {Public\\CNN};

             \draw[thick] (0.45, 2) -- (0.45, 2.7) -- (0.55, 2.7) -- (0.55, 2) -- (0.45, 2)  (0.55, 2) -- (0.85, 2.1) -- (0.85, 2.6) -- (0.55, 2.7);
             \filldraw[thick, fill=my_orange!90] (0.95, 2.1) -- (0.95, 2.6) -- (1, 2.6) -- (1, 2.1) -- (0.95, 2.1) (1, 2.6) -- (1.25, 2.51) -- (1.25, 2.19) -- (1, 2.1);
             \draw[thick, fill=my_orange!90] (1.35, 2.2) -- (1.35, 2.5) -- (1.4, 2.5) -- (1.4, 2.2) -- (1.35, 2.2) (1.4, 2.5) -- (1.55, 2.45) -- (1.55, 2.25) -- (1.4, 2.2);
              \draw[thick] (2.45, 2) -- (2.45, 2.7) -- (2.55, 2.7) -- (2.55, 2) -- (2.45, 2)  (2.55, 2) -- (2.85, 2.1) -- (2.85, 2.6) -- (2.55, 2.7);
             \filldraw[thick, fill=my_orange!90] (2.95, 2.1) -- (2.95, 2.6) -- (3, 2.6) -- (3, 2.1) -- (2.95, 2.1) (3, 2.6) -- (3.25, 2.51) -- (3.25, 2.19) -- (3, 2.1);
             \draw[thick, fill=my_orange!90] (3.35, 2.2) -- (3.35, 2.5) -- (3.4, 2.5) -- (3.4, 2.2) -- (3.35, 2.2) (3.4, 2.5) -- (3.55, 2.45) -- (3.55, 2.25) -- (3.4, 2.2);

             \draw [-stealth] (1, 1.8) -- (2, 1.2);
             \draw [-stealth] (3, 1.8) -- (2, 1.2);

             \filldraw[rounded corners, fill=my_orange!90] (1.25, 0.5) rectangle (2.75, 1.2) node[midway] {Reward};

             \draw (2.75, 0.85) -- (6, 0.85) node[right=-0.4in, above, scale=0.8] {Reward};
             \draw (6, 0.85) -- (6, 3.25);
             \draw[-stealth] (6, 3.25) -- (7, 3.25);
             \draw[-stealth] (8.4, 4.29) -- (8.4, 3.5);

             \node[scale=0.75] at (7.3, 5) {
                \begin{tikzpicture}
                     \filldraw[rounded corners, fill=my_orange!90] (0.25, 1.9) rectangle (1.75, 3.8) node[below left=-0.01in, align=center] {Online Q\\Network};
        
                     \filldraw[rounded corners, fill=my_orange!90] (2.75, 1.9) rectangle (4.25, 3.8) node[below left=-0.01in, align=center] {Target Q\\Network};
        
                     \draw[thick] (0.45, 2) -- (0.45, 2.7) -- (0.55, 2.7) -- (0.55, 2) -- (0.45, 2)  (0.55, 2) -- (0.85, 2.1) -- (0.85, 2.6) -- (0.55, 2.7);
                     \filldraw[thick, fill=my_orange!90] (0.95, 2.1) -- (0.95, 2.6) -- (1, 2.6) -- (1, 2.1) -- (0.95, 2.1) (1, 2.6) -- (1.25, 2.51) -- (1.25, 2.19) -- (1, 2.1);
                     \draw[thick, fill=my_orange!90] (1.35, 2.2) -- (1.35, 2.5) -- (1.4, 2.5) -- (1.4, 2.2) -- (1.35, 2.2) (1.4, 2.5) -- (1.55, 2.45) -- (1.55, 2.25) -- (1.4, 2.2);
                      \draw[thick] (2.95, 2) -- (2.95, 2.7) -- (3.05, 2.7) -- (3.05, 2) -- (2.95, 2)  (2.95, 2) -- (3.35, 2.1) -- (3.35, 2.6) -- (3.05, 2.7);
                     \filldraw[thick, fill=my_orange!90] (3.45, 2.1) -- (3.45, 2.6) -- (3.5, 2.6) -- (3.5, 2.1) -- (3.45, 2.1) (3.5, 2.6) -- (3.75, 2.51) -- (3.75, 2.19) -- (3.5, 2.1);
                     \draw[thick, fill=my_orange!90] (3.85, 2.2) -- (3.85, 2.5) -- (3.9, 2.5) -- (3.9, 2.2) -- (3.85, 2.2) (3.9, 2.5) -- (4.05, 2.45) -- (4.05, 2.25) -- (3.9, 2.2);
        
                \end{tikzpicture}
             };

             \draw (6.1,5.71) -- (6.1,6.95);
             \draw[-stealth] (6.1, 6.95) -- (3.3, 6.95) node[left=-0.6in, above, scale=0.8] {Action};

             \draw[-stealth] (2.75, 5.05) -- (5.8, 5.05) node[right=-0.4in, above, scale=0.8] {State};
             \draw (5, 5.05) -- (5, 4) -- (8.1, 4);
             \draw[-stealth] (8.1, 4) -- (8.1, 4.29);
             
             \filldraw[rounded corners, fill=my_yellow] (7, 3.5) rectangle (8.75, 3) node[midway, scale=0.8] {Target Q};
             \filldraw[rounded corners, fill=my_yellow] (7, 7.6) rectangle (8.75, 7.1) node[midway, scale=0.8] {Current Q};
             \filldraw[rounded corners, fill=my_yellow] (7.2, 6.6) rectangle (8.5, 6.1) node[midway, scale=0.8] {Loss};

             \draw (8.75, 3.25) -- (9.2, 3.25) -- (9.2, 6.35);
             \draw[-stealth] (9.2, 6.35) -- (8.5, 6.35);

             \draw (6.3, 5.71) -- (6.3, 7.35);  
             \draw[-stealth] (6.3, 7.35) -- (7, 7.35);

             \draw[-stealth] (7.85, 7.1) -- (7.85, 6.6);

             \draw[-stealth] (6.93, 5) -- (7.675, 5) node[midway, above, scale=0.6] {Copy};
             \draw [-stealth] (7.2, 6.35) to [bend right=45] (6.5, 5.71);
        \end{tikzpicture}
        };
        \end{tikzpicture}
    \caption{Workflow of our system showing how the quantum camera environment interacts with the DDQN agent. The image data is stored in quantum states, and the agent manipulates these with a set of actions --- quantum circuits that are part of our contribution. Adversarial CNNs create the reward and, during testing, new CNNs are trained from scratch against fixed DDQN weights.}
    \label{fig:blkdiagram}
\end{figure}
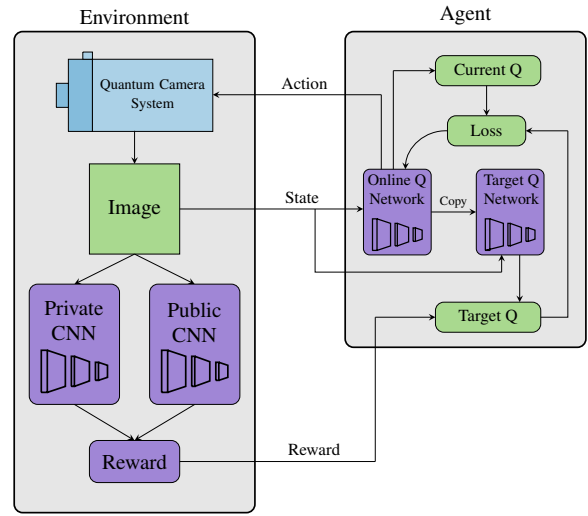

In this paper, we propose a quantum framework where the data, in a quantum sense, is both private and non-private until the point of measurement~\cite{monroe1996schrodinger}. Our design is a hybrid quantum-silicon system, where conventional machine learning sits in silicon, but the data is stored in quantum states. The algorithm learns what actions to take within the quantum computer \emph{before} measurement, such that the measured image has the desired privacy-utility characteristics. 

Our design is unique, and takes advantage of well-known quantum properties that already impact privacy, security and cryptography. One such property is the no-cloning theorem. It states that arbitrary quantum states cannot be perfectly cloned or copied. This is useful for security since it implies an eavesdropper can always be detected\cite{pathakElementsQuantumComputation2013}. In our design, keeping the data in a quantum state provides this level of trust. 

Another property allows lower storage footprints due to quantum computing's unique property of entanglement. With the Flexible Representation of Quantum Images (FRQI), an M by L image can be stored in $n={log}_2\left(ML\right)$ qubits, as opposed to $2^n$ bits on a traditional computer. This means that while the number of traditional bits required to store an image grows exponentially, the number of qubits needed simply grows linearly\cite{yaoQuantumImageProcessing2017}.

\subsection{Contributions}

In this paper, we describe our privacy-preserving design with feasibility experiments. Our work presages the impact of quantum technology on imaging and all of our results are in simulation, due to the limits of technology today. 

Our key idea is to manipulate imagery in quantum states, measuring them as conventional images only when sensitive, private features are protected. However, there is a challenge to manipulating data stored in quantum states. Even a small number of quantum circuits create an exponentially large space of "actions", i.e. image manipulation steps. Additionally, the action space increases with image size. 

We cast the problem of taking the right privacy-preserving actions inside of a quantum framework as a reinforcement learning (RL) task. In this sense, image manipulation is a planning problem, where the goal is a final measured image whose privacy is protected, but many failure states exist along the way. Image measurement at any of these failure states would result in breaching privacy concerns or destroying the public utility. 

Given image data stored in a quantum state, we use an RL-based agent to control the 
quantum privacy preserving camera. The agent's actions provide instructions to the quantum computer, indicating what pixels to redact. The image is then measured and sent to two competing CNNs. The public CNN classifies the image according to the desired non-sensitive classes. The private CNN attempts to classify the image according to the sensitive classes. The reward is generated by rewarding correct public classifications and penalizing correct private classifications, and the agent updates the mask accordingly.

\noindent Our contributions are: 
\setlist{nolistsep}
\begin{enumerate}[noitemsep]
    \item A privacy-preserving design that uses quantum image processing combined with RL-based learning, where the agent's actions are dynamic in training and testing.  
    \item Selection of quantum circuits that would be useful in this privacy preserving scenario. 
    \item Experiments demonstrating that the RL machine learns useful patterns in the large action space, and demonstrating the actual privacy-utility tradeoff by finetuning the CNN classifiers on image data generated by fixed RL policies.
\end{enumerate}

\section{Related Work}

\textbf{Privacy Preservation} When data elements are defined precisely, then formal guarantees for privacy exist~\cite{dwork2008differential}. For imaging, however, such approaches have only found use in image databases, e.g. K-anonymity~\cite{kprivacy}. Privacy-preserving vision has therefore relied on domain knowledge and features such as pixelation, blurring, face/object replacement, etc.---to degrade sensitive information~\cite{fd5,brkic2017know,fd8,fd4,fd2,fd6}. However, we consider the case where adversaries can retrain after our privacy-preserving effect. This makes privacy preservation a harder problem, since deep networks can recover identity from severely degraded data (e.g., \cite{igor}). 
These challenges have opened up research questions at the intersection of inference and privacy\cite{hasan2018viewer}, including methods to avoid capturing sensitive data in the first place\cite{templeman2014placeavoider}. GANs have been utilized to address these challenges. These include using GANs to remove sensitive details from autonomous vehicle data \cite{xiongPrivacyPreservingAutoDrivingGANBased2019}. The GAN approach can have stability issues, however, which the encoder approach helps to alleviate\cite{pittalugaLearningPrivacyPreserving2019}. Another approach beyond GANs is to adaptively sample imagery for robotics. Privacy is maintained by using FPGA or DSP chips to do resolution scaling in hardware, meaning the sensitive data is never stored at a higher resolution. This method is effective but can be difficult to achieve in real-time\cite{kimPrivacyPreservingRobotVision2019}. In our work, we strike out in a complimentary direction of using RL instead of a GAN encoder. While this solves the stability issue, Double Deep Q-learning convergence becomes the deciding factor for performance. 

\textbf{Learning Inside the Camera} Many approaches learn optics and hardware designs to improve post-capture computation accuracy. One example of this strategy can be seen where FGPA and DSP chips are used to scale the resolution of sections of the image pre-capture to achieve privacy\cite{kimPrivacyPreservingRobotVision2019}. Another example uses programmable physical mask to block out light before it reaches the image sensor\cite{pittalugaPreCapturePrivacySmall2017}. Other optical examples use a series of lenses and masks to make an optical convolution layer\cite{changHybridOpticalelectronicConvolutional2018} or natural chromatic aberrations to improve depth perception\cite{changDeepOpticsMonocular2019}. In this paper, the images are in quantum states, and the learned image manipulation happens before measurement. 

\textbf{Quantum Imaging} Currently, noise limits the length of quantum circuits and consequently the size of images. Right now a 2x2 pixel image is the largest FRQI image that can be reliably encoded and measured\cite{geng_improved_2023}. While research is still being done on how to overcome noise in quantum computers, quantum sensors have developed rapidly and shown promising results. For example, sensors have been developed using quantum dot technology to create extremely sensitive covid-19 tests and image sensors\cite{pejovicInfraredColloidalQuantum2022,liSARSCoV2QuantumSensor2022}. The quantum dot image sensors show particular promise in the NIR and SWIR wavelengths\cite{pejovicInfraredColloidalQuantum2022, pejovicPhotodetectorsBasedLead2022,liFlexibleBroadbandImage2018}. Another strategy for realizing quantum cameras involves using photon counting\cite{wolley_quantum_2022}. Smart quantum statistical imaging, using machine learning and photon resolving detection, has been used for superresolving imaging\cite{bhusal_smart_2022}. 

\textbf{Quantum Computing and Security} Quantum computing allows unique security strengths. For example, qubit transfer between two parties cannot be unknowingly intercepted if the parties use a protocol to detect interference, such as quantum key distribution (QKD). Some security challenges quantum computing does face are methods of imperfect cloning and exploitation of hardware imperfections\cite{weierQuantumEavesdroppingInterception2011}. The advantages of quantum computers over classical computers for security outweigh the hazards, and we exploit these in our design.  

\begin{figure}[h]
    \centering
    \includegraphics[width=\linewidth]{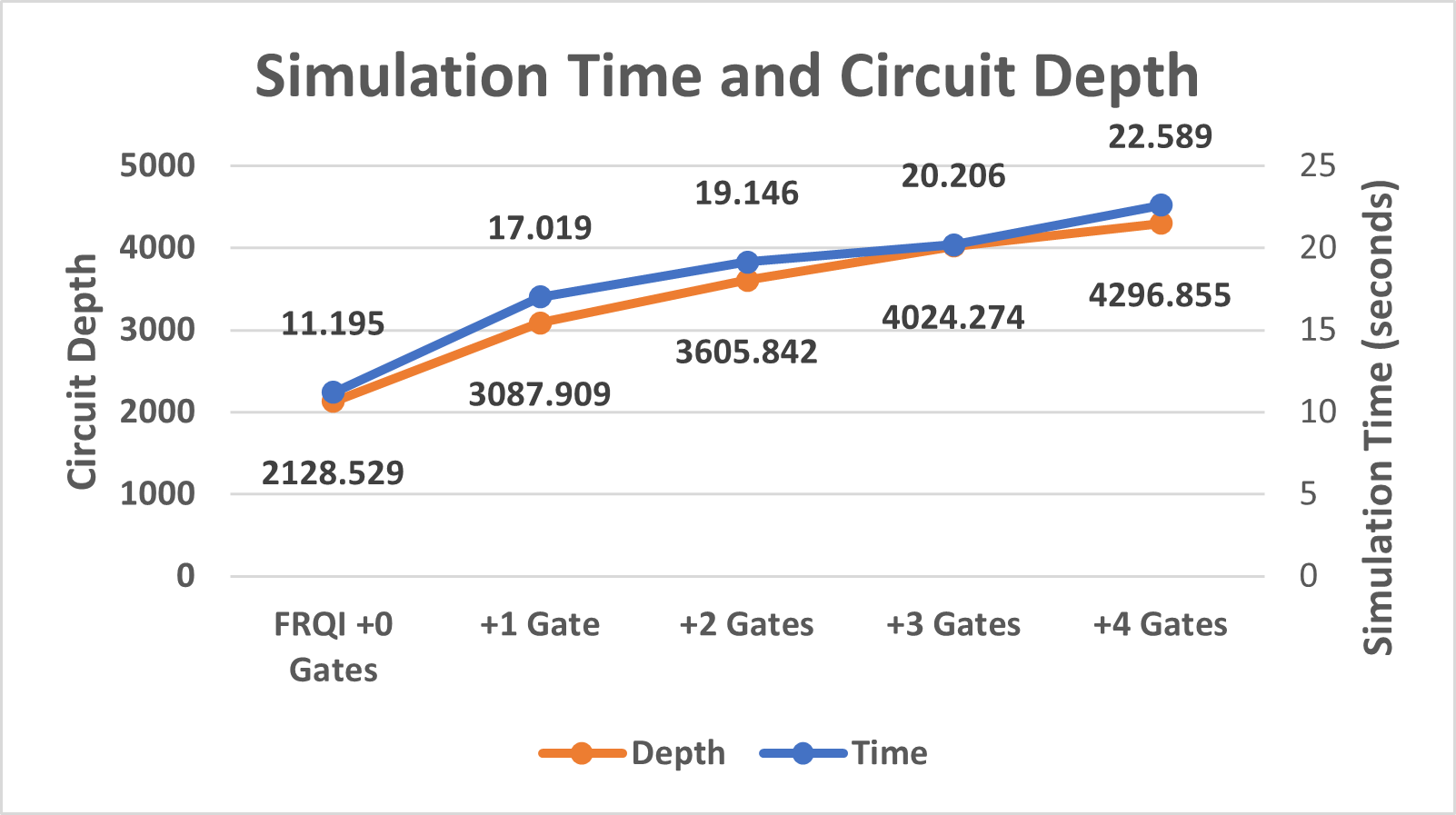}
    \caption{Graph depicting the average simulation time and quantum circuit depth for 16x16 FRQI images. The number of gates on the x-axis refers to the number of additional privacy preserving quantum gates applied after the initial FRQI encoding.}
    \label{fig:circ-depth-time}
\end{figure}

\textbf{Quantum Image Processing} With the development of quantum computing algorithms has come the development of quantum image processing (QImP). In 2003, researchers proposed a method of encoding images in quantum states using a method called Qubit Lattice\cite{yanQuantumComputationBasedImage2014,venegas-andracaStoringProcessingRetrieving2003}. As QImP progressed, other methods were proposed with different pros and cons such as Real Ket\cite{latorreImageCompressionEntanglement2005}, NEQR\cite{zhangNEQRNovelEnhanced2013}, MCQI\cite{yanSurveyQuantumImage2016}, and FRQI\cite{leFlexibleRepresentationQuantum2011,geng_improved_2023}. For this project we chose to focus on FRQI due to its low storage footprint and existing image manipulation algorithms. One advantage of QImP over traditional image processing is that certain image manipulations have exponential speedups with quantum computing such as the Fourier and Hadamard transforms\cite{cavalieriQuantumEdgeDetection2020,yaoQuantumImageProcessing2017}. For example, a quantum edge detection algorithm was developed to run with $O(1)$ processing time rather than the $O(2^n)$ processing time required by classical algorithms\cite{yaoQuantumImageProcessing2017,cavalieriQuantumEdgeDetection2020}. Additionally, security methods such as watermarking are already in existence\cite{yanQuantumComputationBasedImage2014}. 

One limitation of FRQI is that it stochastically encodes images into quantum states. The accuracy of the pixel grey scale value is directly related to how many times the circuit is run (often referred to as the number of “shots”)\cite{leFlexibleRepresentationQuantum2011,geng_improved_2023}. This phenomenon is described by the no-teleportation theorem, which states that it is impossible to exactly reconstruct a quantum state with classical bits. In this paper, we convert this ``bug" into a ``feature" and use such quantum circuits to our advantage for privacy applications.

\textbf{Flexible Representation of Quantum Images} The Flexible Representation of Quantum Images (FRQI) is a method of representing images as a quantum state. For a $2^n\times2^n$ pixel image, $2n+1$ qubits are needed to encode the image in a quantum state using FRQI. The location of each pixel is stored in the $2n$ qubits while the color information is stored in a single qubit as a scaled value between $0$ and $\frac{\pi}{2}$. Using the computational basis $|1\rangle$ and $|0\rangle$, the FRQI image is encoded in a quantum state by:
\[|I(\theta)\rangle=\frac{1}{2^n}\sum^{2^{2n}-1}_{i=0}(\cos\theta_i|0\rangle+\sin\theta_i|1\rangle)\otimes|i\rangle,\]
\[\theta_i\in[0,\frac{\pi}{2}], i=0,1,\dots,2^{2n}-1\]
Where $\theta_i$ is the color information for each pixel location $i$, and $n=\log_2(\text{image length})$\cite{leFlexibleRepresentationQuantum2011,geng_improved_2023}. 

\begin{figure}[h]
    \centering
        \begin{tikzpicture}
            \filldraw[fill=black!70, draw=black] (0,0) rectangle (2,2) node[midway, white] {$|11\rangle$, $\frac{\pi}{51}$};
            \filldraw[fill=black!00, draw=black] (0,2) rectangle (2,4) node[midway] {$|00\rangle$, $\frac{\pi}{2}$};
            \filldraw[fill=black!100, draw=black] (2,0) rectangle (4,2) node[midway, white] {$|11\rangle$, $0$};
            \filldraw[fill=black!30, draw=black] (2,2) rectangle (4,4) node[midway] {$|00\rangle$, $\frac{\pi}{3}$};
        \end{tikzpicture}
    \caption{2x2 FRQI image, showing the $|i\rangle$ location information and $\theta_i$ color information}
    \label{fig:frqi}
\end{figure}
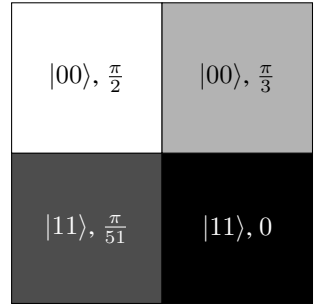

\begin{figure}[h]
    \centering
    \begin{tabular}{cc}
    \includegraphics[width=0.27\linewidth]{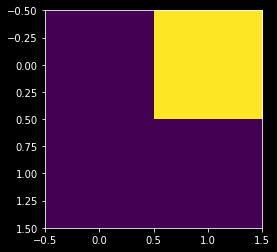} & \includegraphics[width=0.27\linewidth]{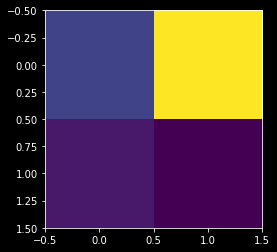}\\
    Original Input Image & Measured FRQI Image\\
    \end{tabular}
    \centering
    \begin{tabular}{ccc}
    \includegraphics[width=0.27\linewidth]{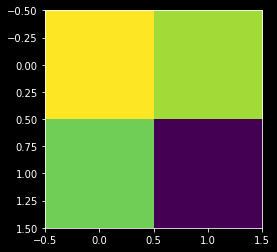} & \includegraphics[width=0.27\linewidth]{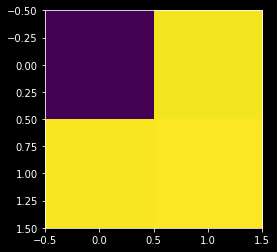} & \includegraphics[width=0.27\linewidth]{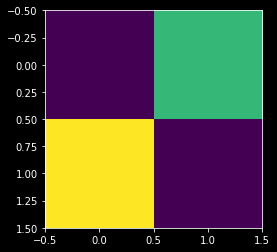} \\
    + Gate 1 & + Gate 2 & + Gates 1 and 2\\
    \end{tabular}
    \caption{\textbf{Real} low-resolution FRQI quantum images run on `ibmq\_manila`, an IBM Quantum Falcon processor\cite{ibm_ibm_2022}. The first image is the classical input image. The second is the measured FRQI image with no additional gates applied. Gate 1 is the measured FRQI image with an additional gate added (Controlled-RX on $p0, c$). Gate 2 is the measured FRQI image with a different additional gate added (Controlled-RX on $p1, c$). Gates 1 and 2 is the measured FRQI image with both gates 1 and 2 sequentially applied to the original FRQI image.}
    \label{fig:real-bckend}
\end{figure}

\section{Experimental Challenges} 
We faced challenges due to the computational complexity of quantum simulations and circuit size constraint on current NISQ (noisy intermediate-scale quantum era) quantum computers. A 2x2 pixel image is the largest FRQI image that can be reliably recovered from current quantum computers due to noise that arises with longer quantum circuit depth\cite{geng_improved_2023}. Using Qiskit’s AerSimulator, the largest reliably simulated FRQI image is 16x16 pixels\cite{geng_improved_2023}. 

Additionally, simulation latency to prepare and run the quantum circuits is a challenge. Fig. \ref{fig:circ-depth-time} shows that the average time to simulate a 16x16 FRQI image with a NVIDIA GeForce GTX 1080 Ti is 11 seconds. The simulation time increases with quantum circuit depth, so that four additional gates after the initial FRQI encoding effectively doubles the simulation time per image. Working within these constraints, rescaled 16x16 EMNIST images were used as the data set with the public utility classes being whether an image represents a ‘letter’ or a ‘number’ and the private, sensitive classes being the exact letter or number (the uppercase letters and numbers from the balanced EMNIST dataset\cite{cohen_emnist_2017}).

\textbf{Experimental Proof of Concept} Our proposed solution involves the combination of quantum gates applied to an FRQI image to achieve privacy preservation. While 16x16 images are used in our simulation results, providing more options for potential privacy preservation through quantum gates, the same concept can still be seen in 2x2 images. In Fig. \ref{fig:real-bckend}, a 2x2 FRQI image is run on ‘ibmq\_manila,’ which is a 5 qubit IBM Quantum Falcon processor.  The input image is compared to the result with no additional gates after FRQI encoding, one controlled-RX gate (applied to either positional qubit), or two controlled-RX gates (applied to both positional qubits). This shows the different outputs that can be achieved with one type of quantum gate, and how they interact together to produce different results in combination.

\section{Quantum Privacy Preserving System}

In Fig. \ref{fig:blkdiagram}, we show our work-flow, where data from the scene is captured and stored in quantum states. This represents the environment in our reinforcement learning setup, whereas the agent attempts to manipulate the quantum states by controlling internal quantum circuits. We discuss the environment and the agent in detail below.

\subsection{Quantum States as an Environment}

\begin{figure}[h]
    \centering
    \begin{tabular}{P{3.0cm}P{3.0cm}}
    \large{\textbf{No Changes}} & \large{\textbf{Translation}} \\
    \includegraphics[width=3cm]{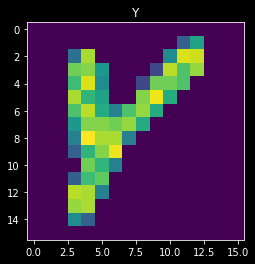} & \includegraphics[width=3cm]{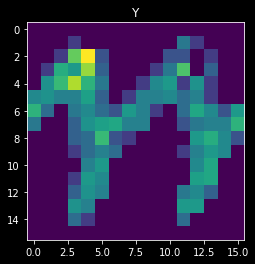} \\
    Controlled-Z Gate ($p_3, c$) & Hadamard Gate\break ($p3$) \\
    \includegraphics[width=3cm]{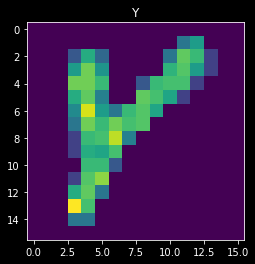} & \includegraphics[width=3cm]{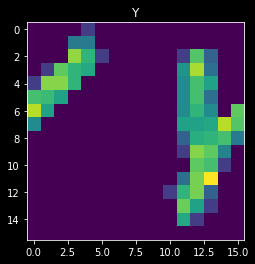} \\
    Controlled-RZ Gate ($p_3, c, \frac{\pi}{2}$) & X Gate\break ($p3$) \\
    & \cellcolor{my_blue!60!white}\\
    \large{\textbf{Inversion}} & \cellcolor{my_blue!60!white}\Large{\textbf{Complex}} \\
    \includegraphics[width=3cm]{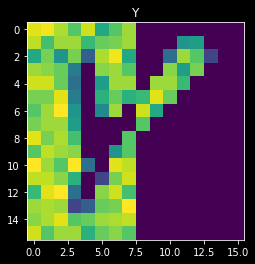} & \cellcolor{my_blue!60!white}\includegraphics[width=3cm]{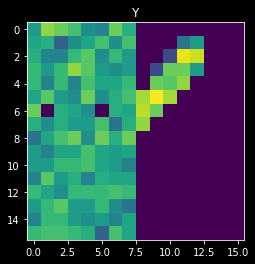} \\
    Controlled-X Gate ($p_3, c$) &\cellcolor{my_blue!60!white} Controlled-RX Gate ($p_3, c, \frac{\pi}{2}$) \\
    \includegraphics[width=3cm]{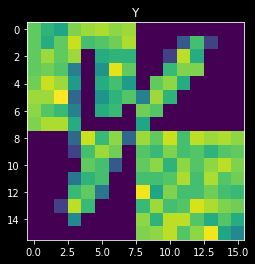} & \cellcolor{my_blue!60!white}\includegraphics[width=3cm]{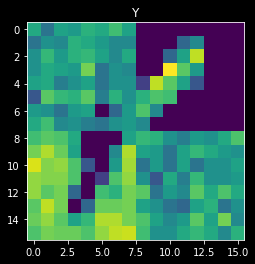} \\
    Controlled-X Gates ($p_3, p_7, c$) & \cellcolor{my_blue!60!white}Controlled-RX Gates ($p_3, p_7, c, \frac{\pi}{2}$) \\
    & \cellcolor{my_blue!60!white}\\
    \end{tabular}
    \caption{Many quantum gates, such as Controlled-Z gate, do not affect images. Others, such as the Hadamard Gate and the controlled-X gate demonstrate simple transformations. We have selected Controlled-RX gates that can provide sophisticated image manipulation.}
    \label{fig:gates-examples}
\end{figure}

\begin{figure*}[h]
    \centering
        \begin{tikzpicture}
        \fill[left color=my_blue!30!white, right color=my_blue!70!white, shading angle=90] (-5.3,-1) -- (-5.3,1) -- (-4.5, 1) -- (-4.5, 1.5) -- (-3.5,0) -- (-3.5,0) -- (-4.5, -1.5) -- (-4.5, -1) -- (-5.3, -1);
        \node at (-7,0) {\includegraphics[width=3cm]{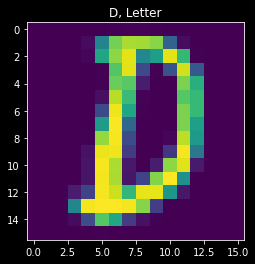}};

          \node[scale=0.7] at (0,0) {
            \begin{quantikz}[row sep={0.6cm,between origins}]
                \lstick{$c$}   & \gate[wires=9]{\begin{array}{c} \text{FRQI} \\ \text{Encoder} \end{array}} & \gate{R_x(\frac{\pi}{2})}  & \qw & \gate{R_x(\frac{\pi}{2})} & \gate{R_x(\frac{\pi}{2})} & \qw\\
                \lstick{$p_0$} & &\ctrl{-1} & \qw & \qw & \qw & \qw \\
                \lstick{$p_1$} & &\qw  & \qw & \qw & \qw & \qw \\
                \lstick{$p_2$} & &\qw  & \qw & \qw & \qw  & \qw\\
                \lstick{$p_3$} & &\qw  & \qw & \qw & \qw  & \qw\\
                \lstick{$p_4$} & &\qw  & \qw & \ctrl{-5} & \qw & \qw \\
                \lstick{$p_5$} & &\qw  & \qw & \qw & \qw  & \qw\\
                \lstick{$p_6$} & &\qw  & \qw & \qw & \qw  & \qw\\ 
                \lstick{$p_7$} & &\qw  & \qw & \qw & \ctrl{-8}  & \qw
            \end{quantikz}
           };
        \fill[left color=my_blue!30!white, right color=my_blue!70!white, shading angle=90] (3.5,-1) -- (3.5,1) -- (4.3, 1) -- (4.3, 1.5) -- (5.3,0) -- (5.3,0) -- (4.3, -1.5) -- (4.3, -1) -- (3.5, -1);
        \node at (7,0) {\includegraphics[width=3cm]{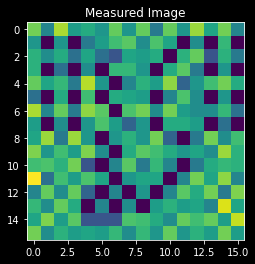}};
        \end{tikzpicture}

        \centering
        \begin{tabular}{P{3.0cm}P{3.0cm}P{3.0cm}P{3.0cm}P{3.0cm}}
            \toprule
            {\small Original Character} & {\small $\text{Positional Qubits: }\allowbreak p_3, p_7$} & {\small $\text{Positional Qubits: }\allowbreak p_0, p_4, p_7$} & {\small $\text{Positional Qubits: }\allowbreak p_2, p_3, p_5, p_7$} & {\small $\text{Positional Qubits: }\allowbreak p_2, p_3, p_4, p_6, p_7$} \\
            \midrule
            \includegraphics[width=3cm]{d.png} & \includegraphics[width=3cm]{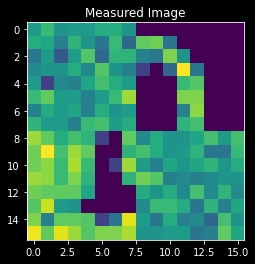} & \includegraphics[width=3cm]{d_0_4_7.png} & \includegraphics[width=3cm]{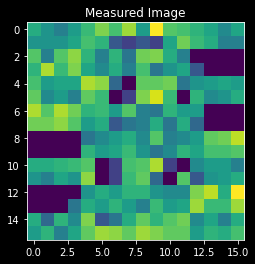} & \includegraphics[width=3cm]{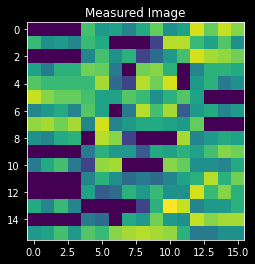} \\
            \includegraphics[width=3cm]{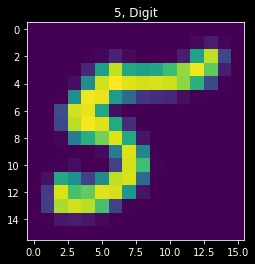} & \includegraphics[width=3cm]{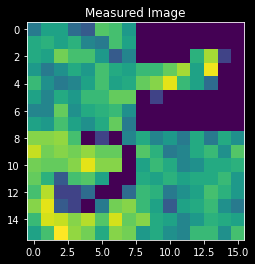} & \includegraphics[width=3cm]{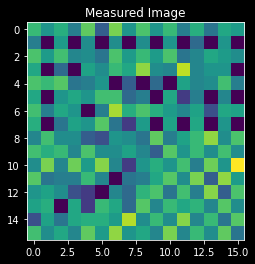} & \includegraphics[width=3cm]{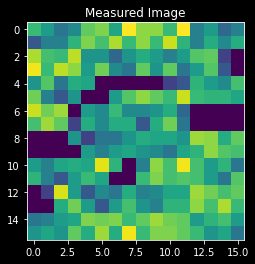} & \includegraphics[width=3cm]{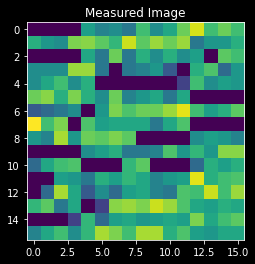} \\
            \bottomrule
        \end{tabular}

        \captionof{figure}{We built a FRQI encoder out of the quantum gates that we selected for privacy-preservation. The $16 \times 16$ images are stored in 9 qubits. This corresponds to the efficiency of storing a $2^4\times2^4$ pixel image in $2*4+1$ qubits. Manipulating images in lower dimensional qubits is non intuitive and, further, the size of the action space is 24,157. The action space produces interesting manipulations are shown in the figure for a small sampling of actions (four sets of actions for the letter D and number 5). We learn the best actions for a particular privacy preserving task.         }
        \label{fig:big-gates-output}
\end{figure*}

\textbf{Privacy through Quantum Gates} For this project, we explored different methods of redacting or destroying FRQI image information. The privacy for these methods is found after measurement, as quantum computations are inherently reversible. Therefore, we treat the quantum circuit like a black box, using gates and controls to manipulate the unknown image before measuring the result.

We have explored the space of quantum circuits and selected those that are useful for privacy preservation. A first approach for destroying FRQI image information is by controlling the number of shots, analogous to photon noise in image capture. This comes at the advantage of making the overall computation faster (as the circuit is run fewer times to obtain a noisy image) but at the disadvantage of not having precise control. 

The second method is by adding gates to the FRQI image to redact parts of the initial image. Some manipulations can be done with one gate, redacting chunks of the image. Similar methods have been used to translate images or do color manipulations\cite{yanQuantumComputationBasedImage2014,iliyasu_towards_2013}. 

The gates perform color transformations on select patterns of the image. The patterns are selected by which positional qubits are used as the controls for a controlled X rotation. The rotation angle is selected as $\theta=\frac{\pi}{2}$ to obscure image information in the selected areas. 

\emph{Selection of these gates was not trivial.} Fig. \ref{fig:gates-examples} shows selected examples of manipulations that can be achieved with minimal gates. Only controlled quantum rotations were found to be the most promising for our use case since information is effectively redacted rather than simply translated or inverted. Due to the outputs being similar between controlled-RX and controlled-RY gates, controlled-RX gates were selected to be used in our action set.

Of course, one could follow the approach of many QImP techniques and redact single pixels using the FRQI equation. This comes at the cost of being exponentially more computationally expensive compared to the former two methods, but it allows more granular control. Adding noise can also be achieved this way by randomly picking a number of pixels to redact.

There is an exponentially large number of potentially privacy preserving actions that can be taken on an FRQI image. We limited the action space by allowing up to 4 of a total of 28 actions to be selected for a single image. These actions consist of single controlled-RX gates and levels of noise. This limited space already consists of 24,157 unique actions that can be taken on an FRQI agent. If image size, allowed action type, or number of gates is increased this action space size would increase further. Our action space is therefore designed to allow more varied and rich image outputs while limiting the overall size of the action space. Fig. \ref{fig:big-gates-output} shows a few examples of the possible outputs that can be achieved by combining these gates.

\begin{figure*}[h]
    \centering
    \begin{tabular}{P{0.3\textwidth}P{0.3\textwidth}P{0.3\textwidth}}
        \textbf{Q Online and Q Target Values} & \textbf{Mean Loss} & \textbf{Smoothed Reward} \\
        \\
        \includegraphics[width=\linewidth]{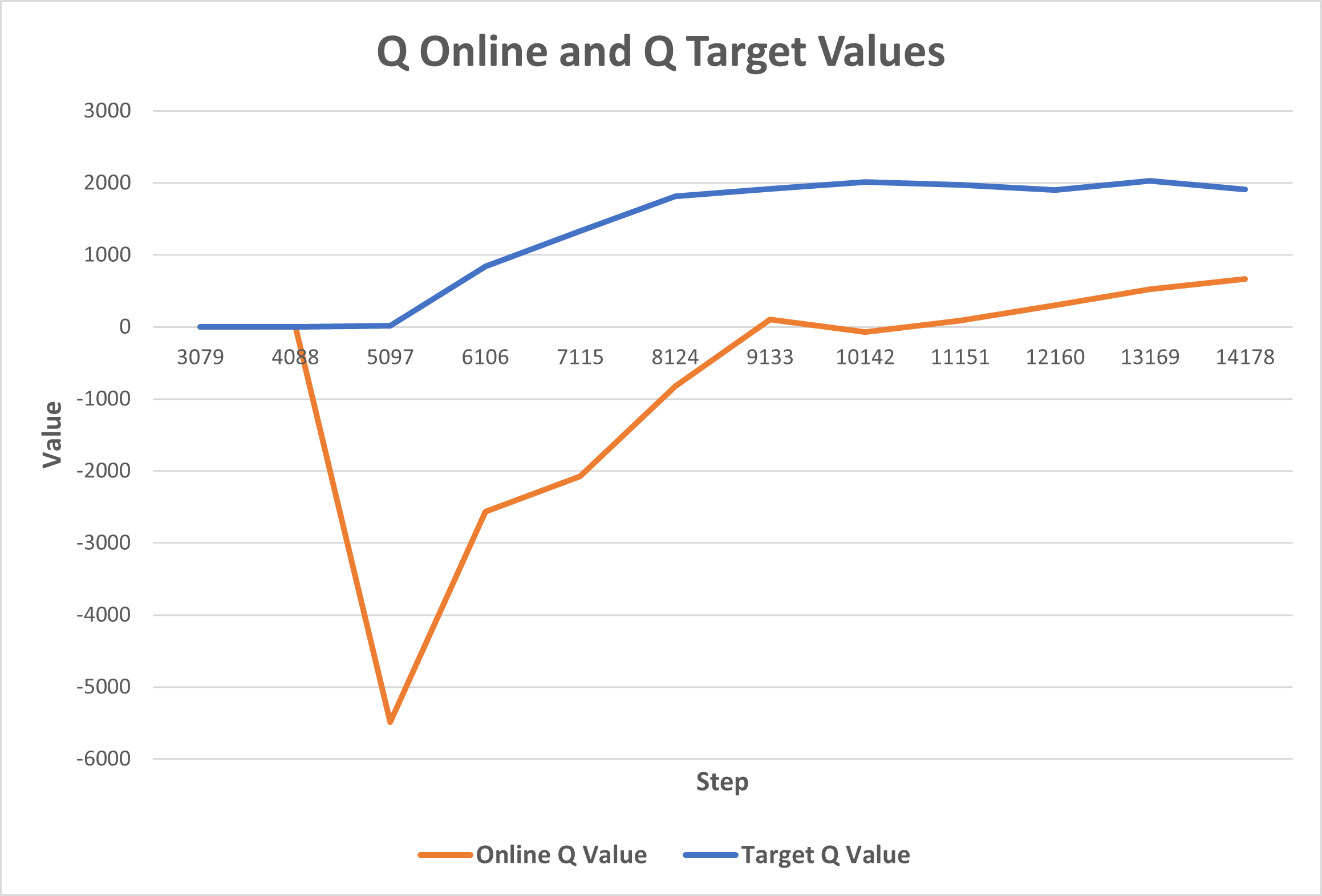} \footnotesize{(a) Public-Based Reward Policy} \label{fig:q-graph-length} &  \includegraphics[width=\linewidth]{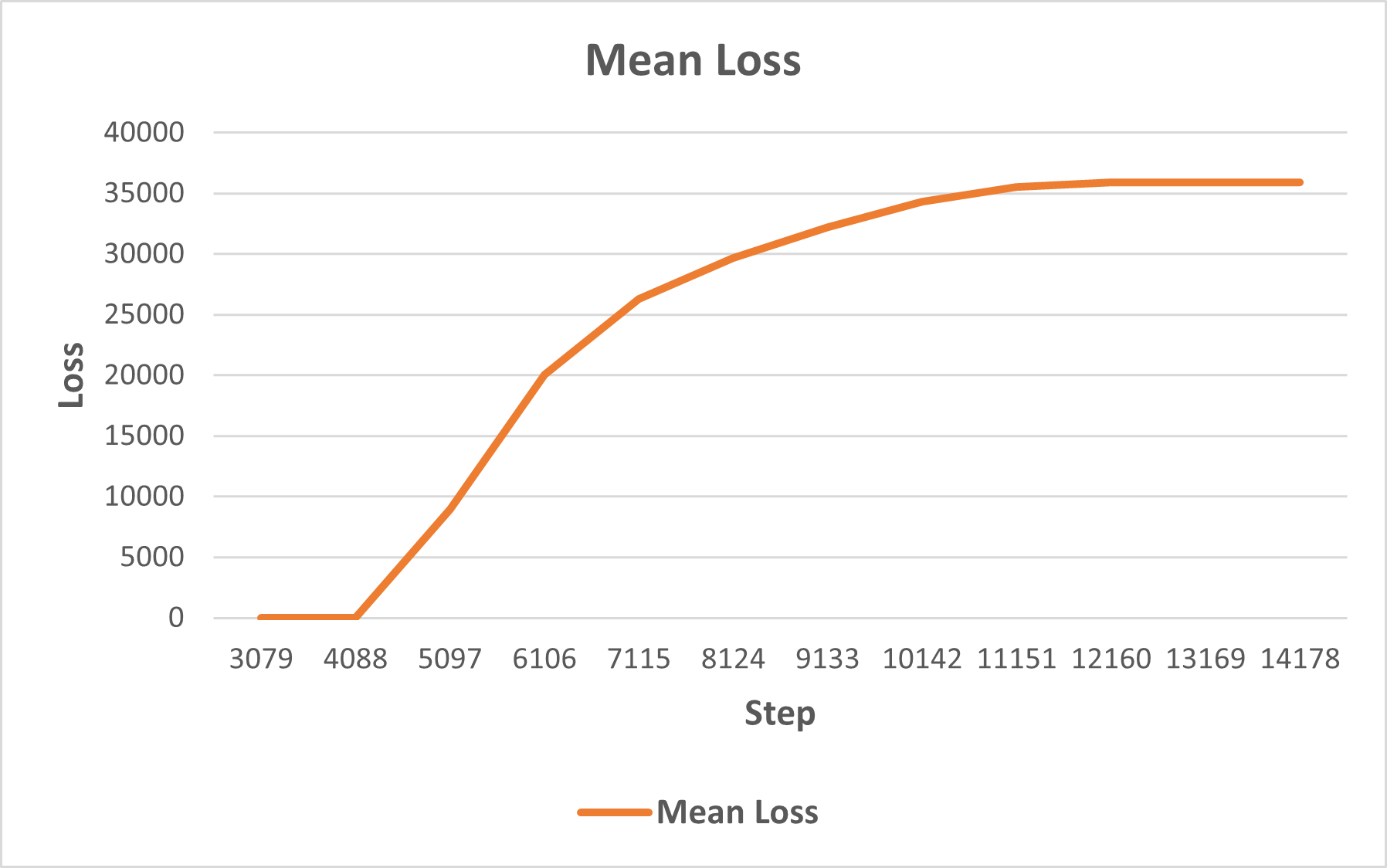} \footnotesize{(b)} \label{fig:loss-length} & \includegraphics[width=\linewidth]{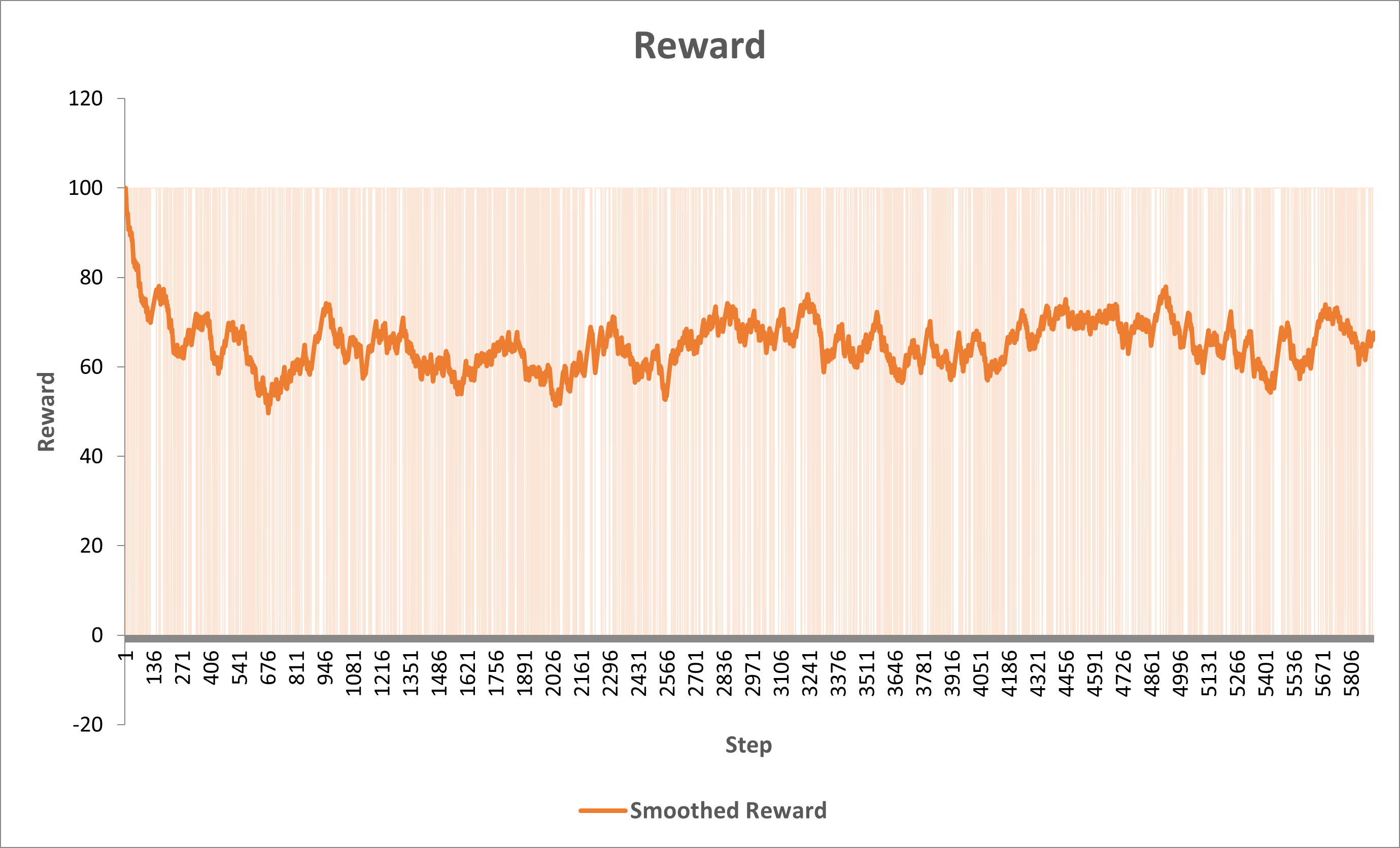} \footnotesize{(c)}\label{fig:reward-length} \\
        \includegraphics[width=\linewidth]{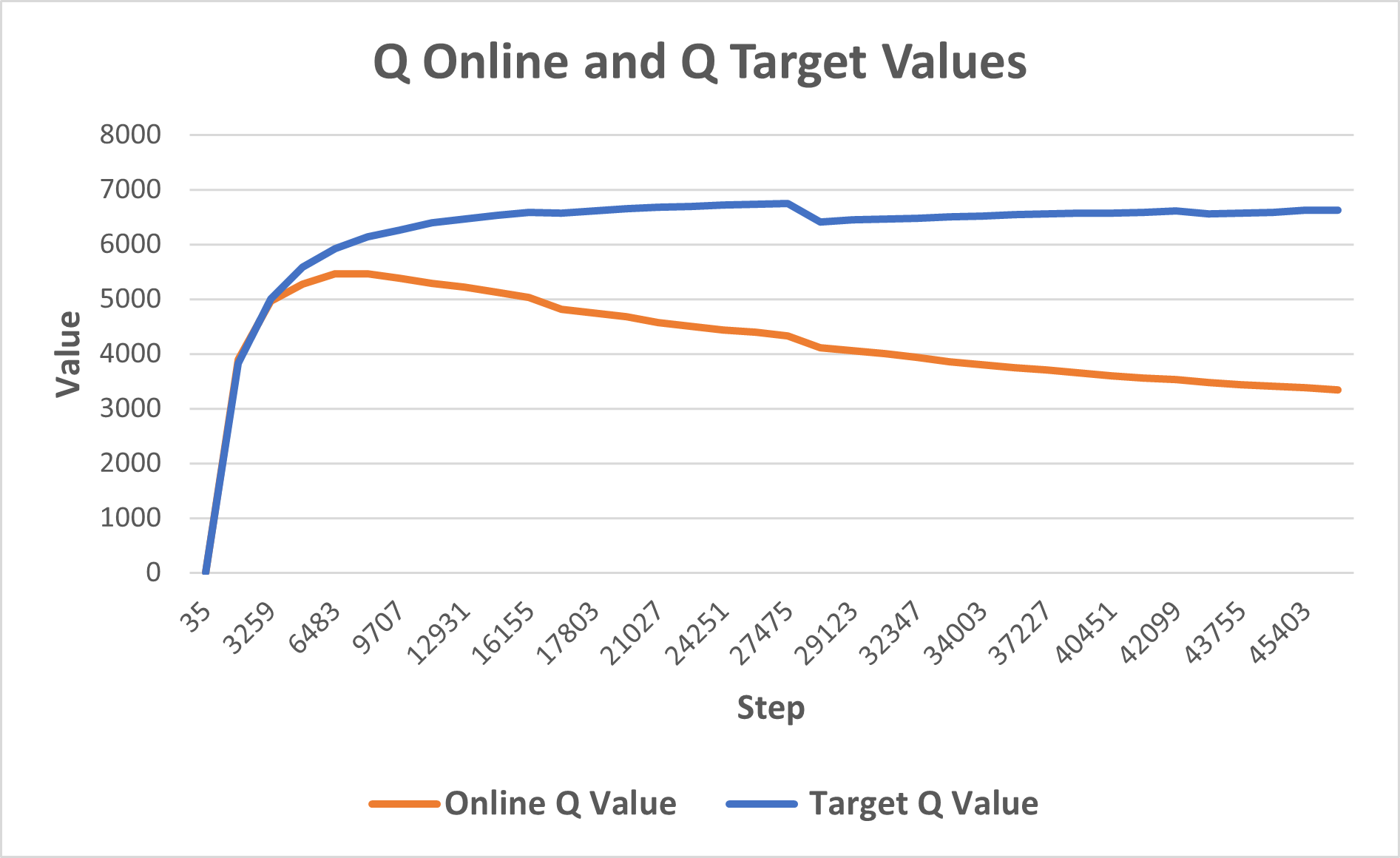} \footnotesize{(a) Public-Private Reward Policy} \label{fig:q-graph-1} & \includegraphics[width=\linewidth]{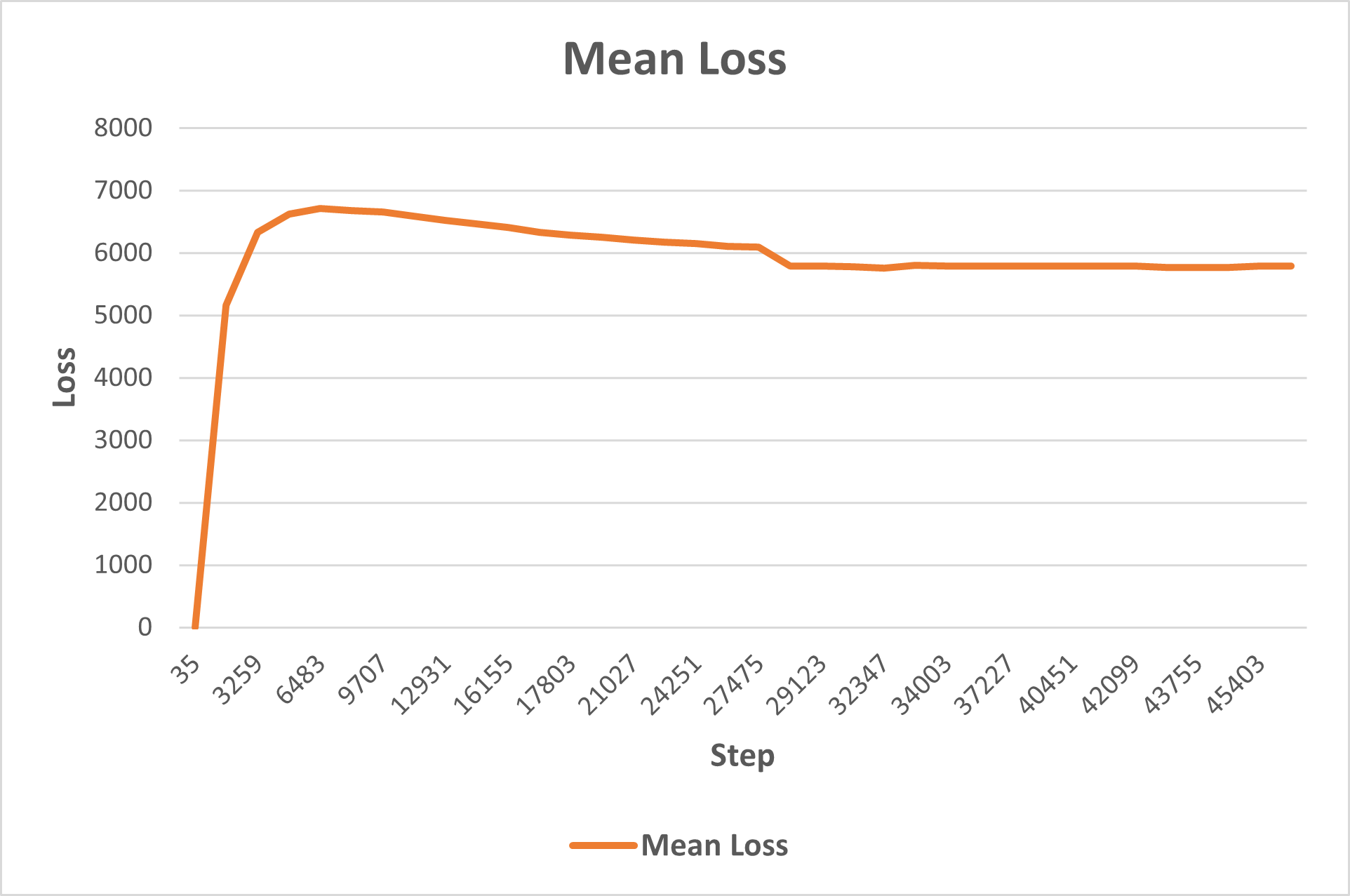} \footnotesize{(b)}\label{fig:loss-1} & \includegraphics[width=\linewidth]{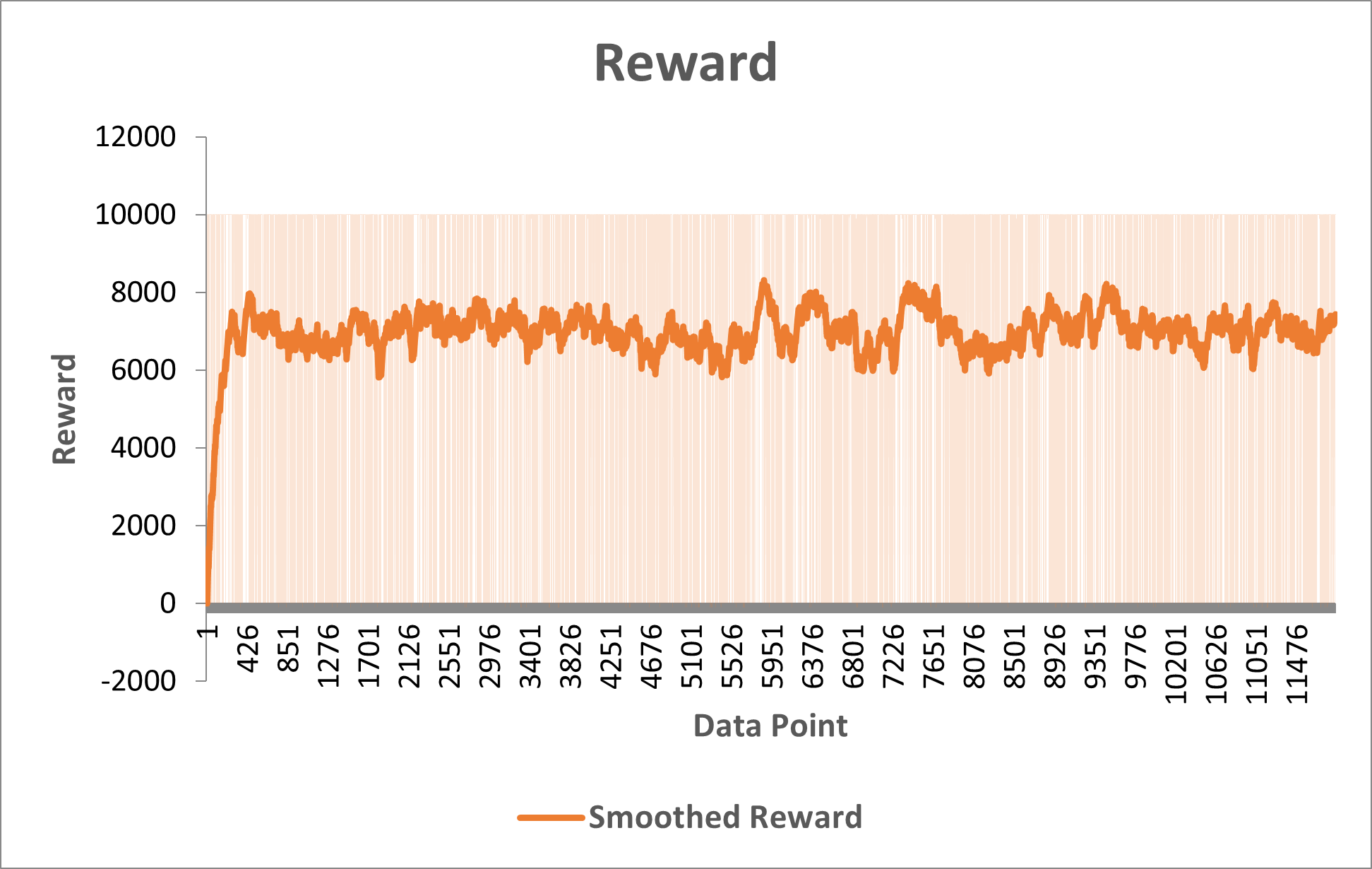} \footnotesize{(c)}\label{fig:reward-1} \\
        \includegraphics[width=\linewidth]{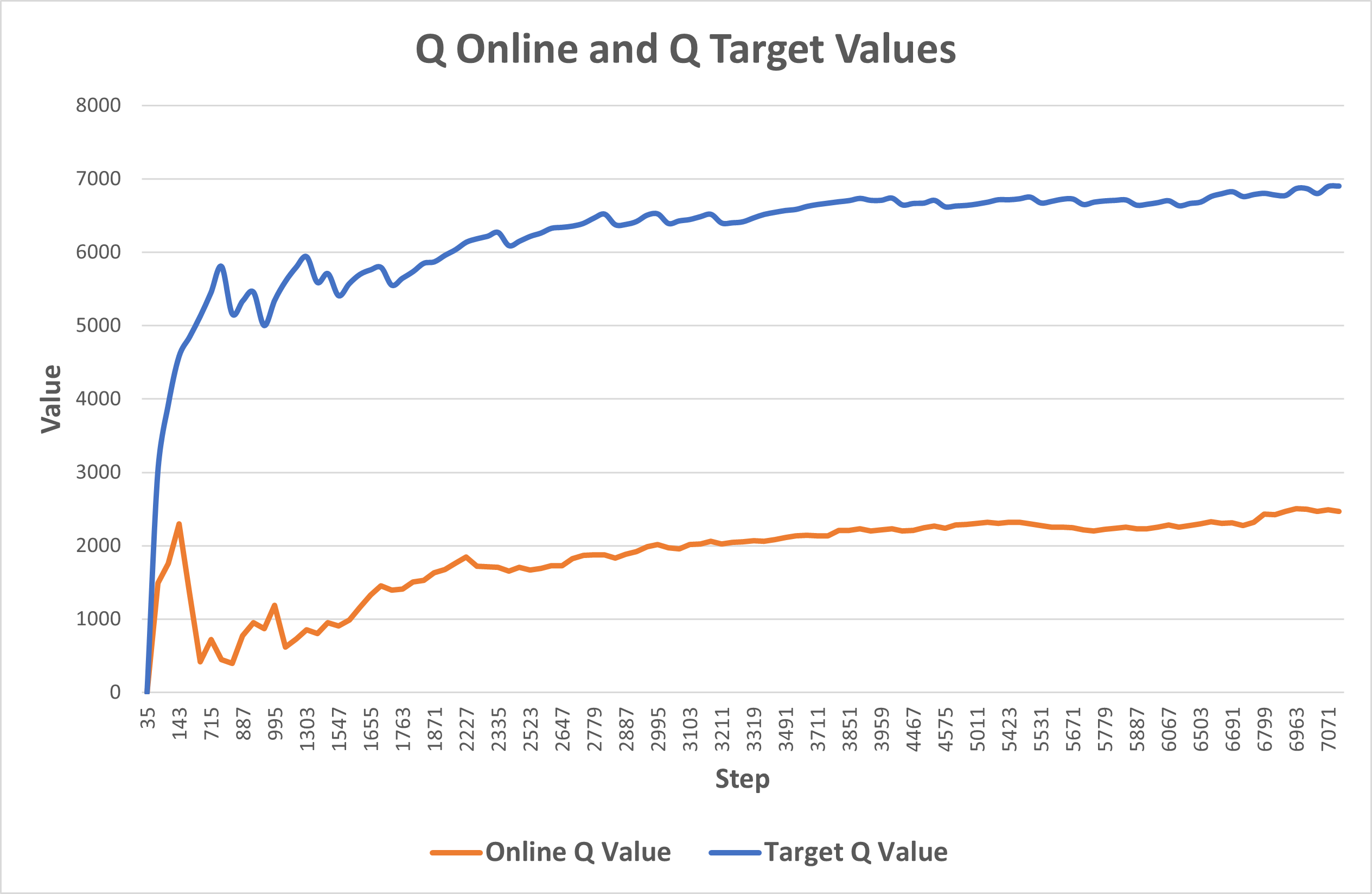} \footnotesize{(a) Length-Based Reward Policy}\label{fig:q-graph-2} & \includegraphics[width=\linewidth]{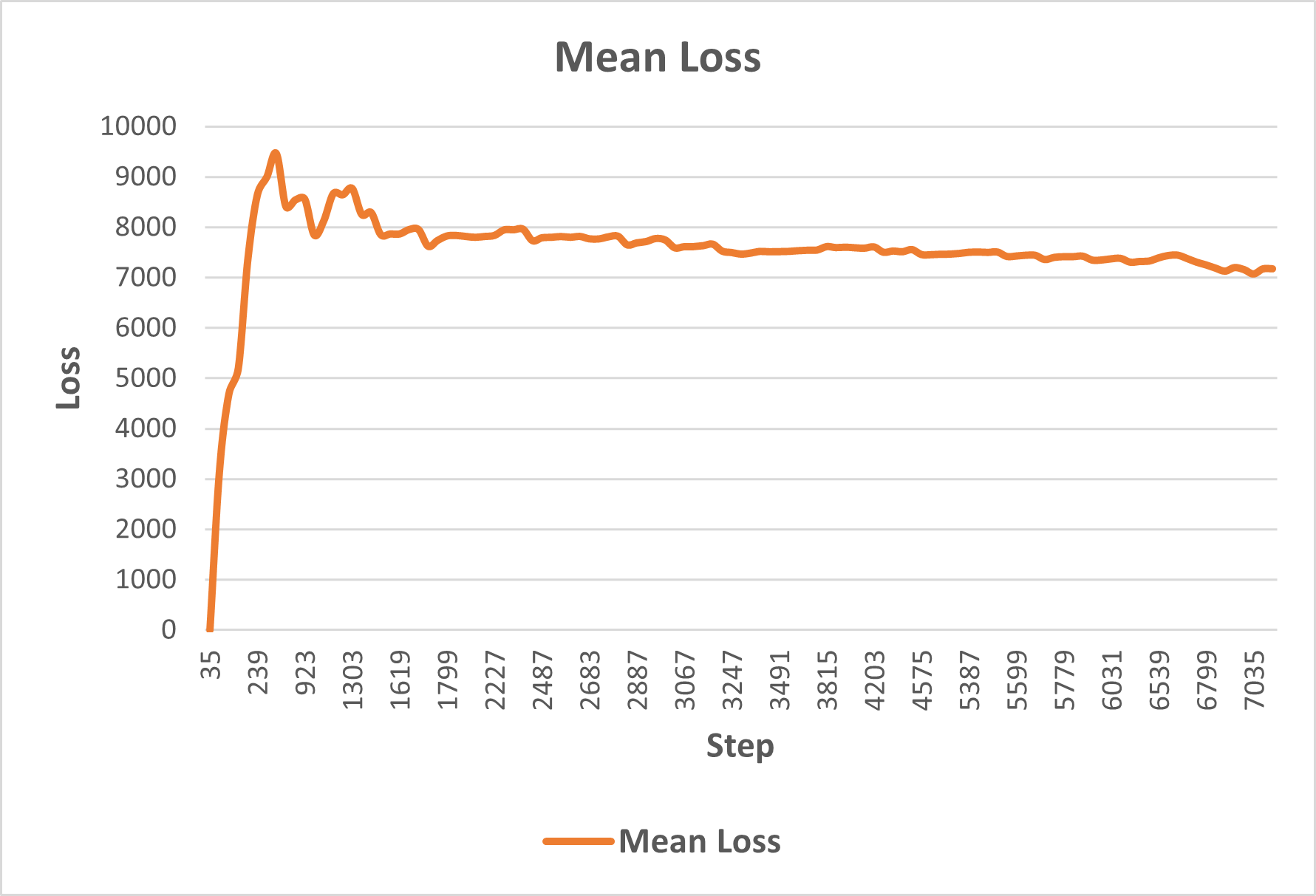} \footnotesize{(b)}\label{fig:loss-2} & \includegraphics[width=\linewidth]{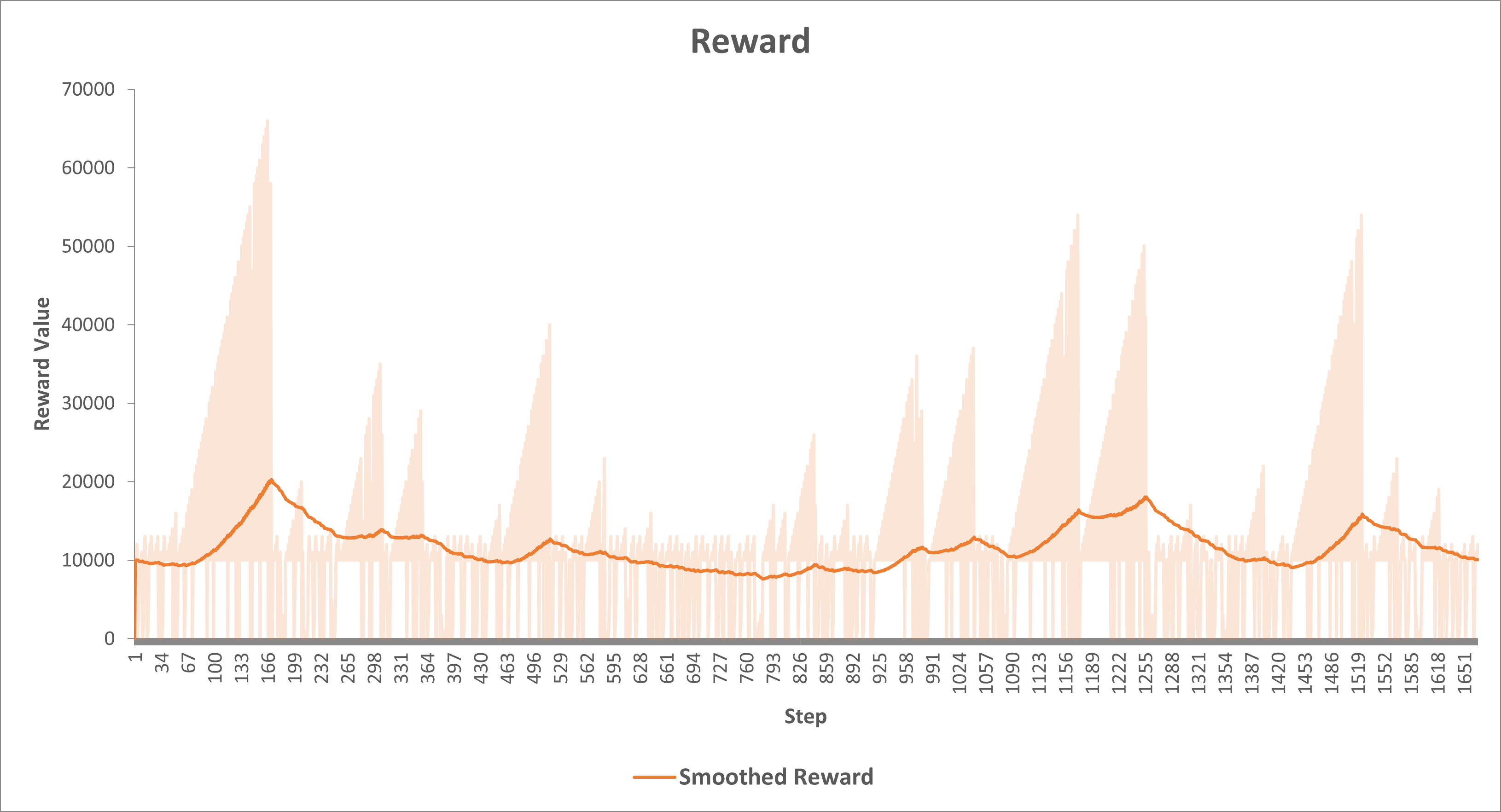} \footnotesize{(c)}\label{fig:reward-2} \\
        \includegraphics[width=\linewidth]{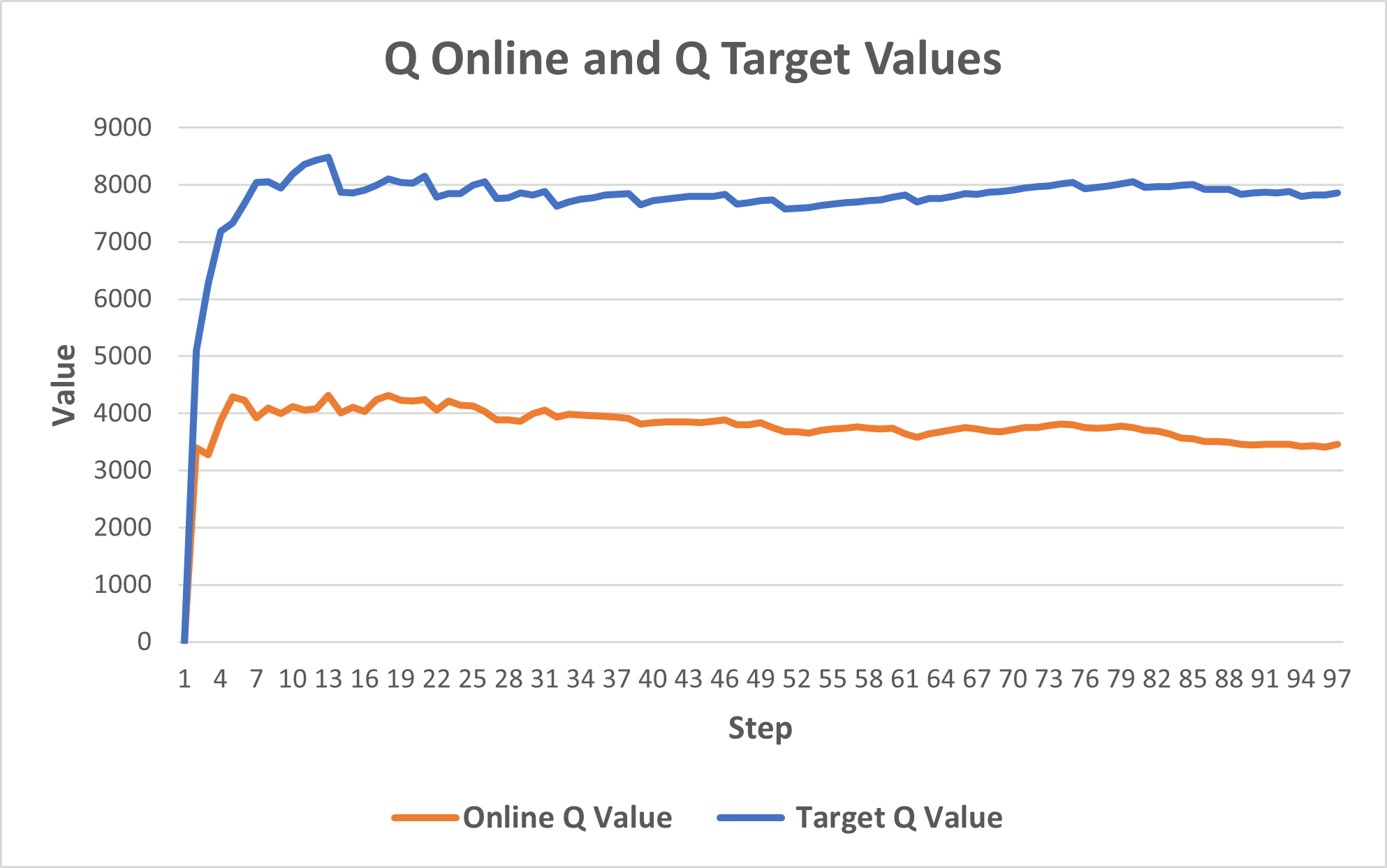} \footnotesize{(a) Accuracy-Based Reward Policy}\label{fig:q-graph-3} & \includegraphics[width=\linewidth]{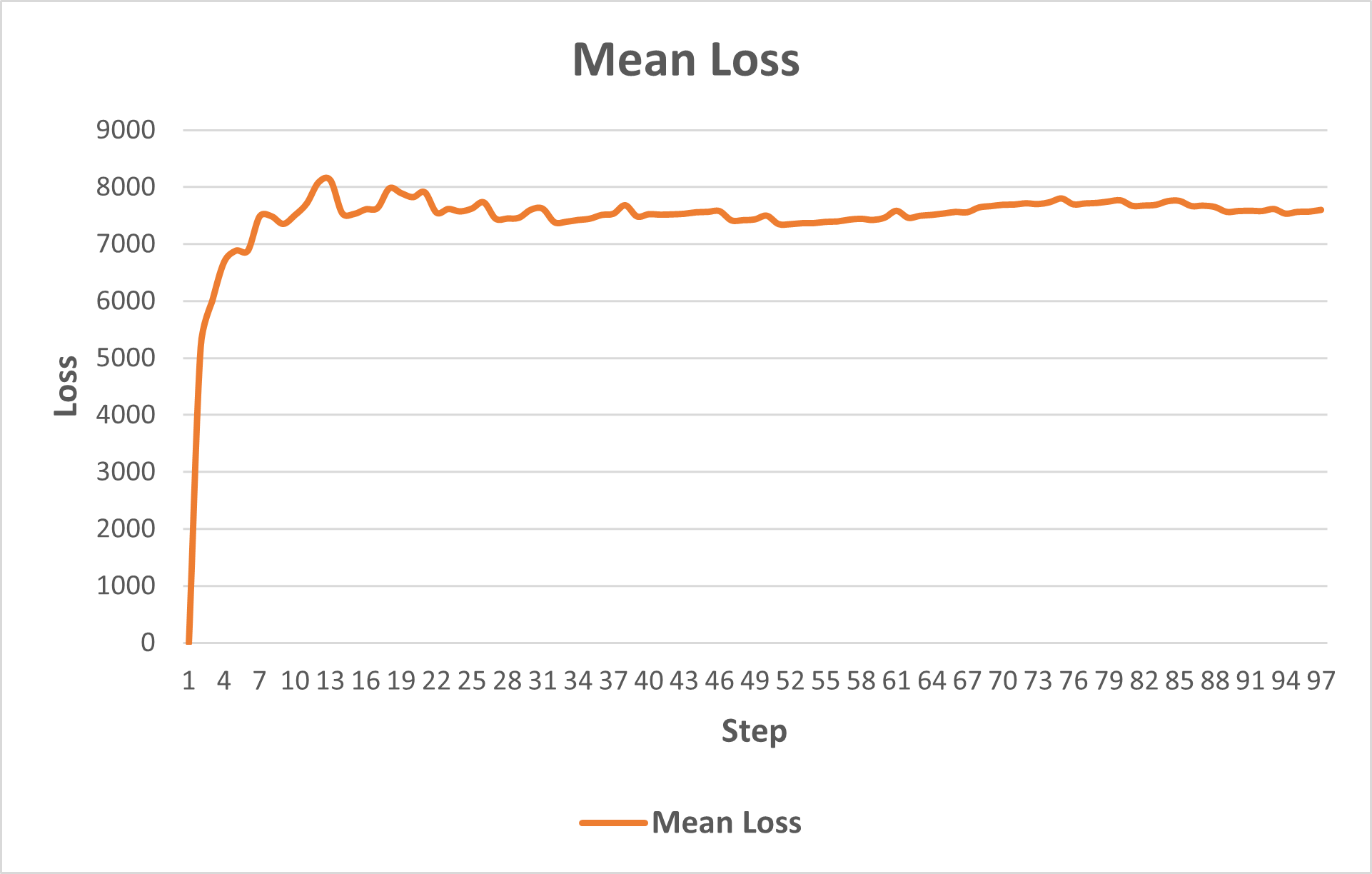} \footnotesize{(b)}\label{fig:loss-3} & \includegraphics[width=\linewidth]{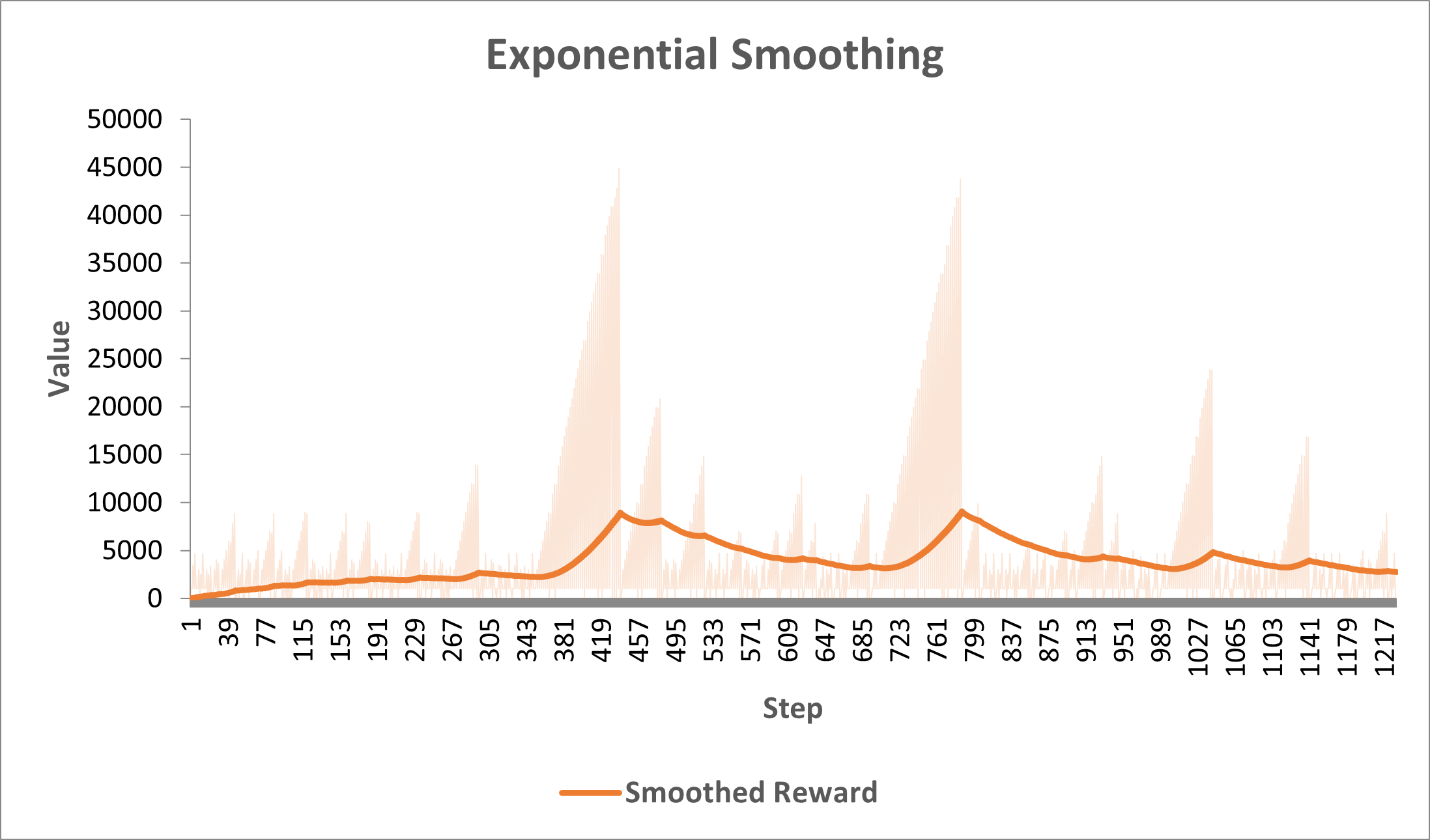} \footnotesize{(c)}\label{fig:reward-3} \\
    \end{tabular}
    \caption{Here we show the DDQN training graphs for four rewards that we investigate, with one in each row. For each experiment, the convergence of the Q values and the stability of the loss show the machine is learning. The results of these are shown in Table \ref{fig:results-table}.}
    \label{fig:graph-table}
\end{figure*}

\begin{figure*}[h!]
    \centering
    \begin{tabular}{|P{4.5cm}||P{3.45cm}|P{3.43cm}|P{2.85cm}|}
      \hline
      \multicolumn{4}{| c |}{\textbf{Privacy and Utility Preservation Performance}}\\
      \hline
      \textbf{Pure Chance Baseline} & {\textbf{50.0\%}} &{\textbf{2.8\%}} & {}\\
      \hline \hline
      \textbf{Data} & {\textbf{Public CNN Accuracy}} & \textbf{Private~CNN~Accuracy} & {}\\
      \hline
      Quantum Augmentation Baseline & {88.5\%} & {85.0\%} & {}\\
      \hline
    \end{tabular}
    \begin{tabular}{|P{4.5cm}||P{1.35cm}|P{1.65cm}|P{1.35cm}|P{1.65cm}|P{1.2cm}|P{1.2cm}|}
      \hline
      & \textbf{Testing} & \textbf{Finetuned} & \textbf{Testing} & \textbf{Finetuned} & \textbf{Training Steps} & \textbf{Training Hours}\\
      \hline
      Quantum Random Action Baseline & 49.6\% & 51.9\% & 2.7\% & 2.1\% & -- & -- \\
      Gaussian Blur Augmented EMNIST Baseline & 54.1\% & 83.9\% & 6.1\% & 80.7\% & -- & -- \\
      Gaussian Noise Augmented EMNIST Baseline & 51.5\% & 56.5\% & 3.3\% & 11.4\% & -- & -- \\
      \hline \hline
      \textbf{Public-Based Reward Policy} & \textbf{50.7\%} & \textbf{56.5\%} & \textbf{4.6\%} & \textbf{2.5\% } & 90,397 & 605\\
      Length-Based Reward Policy & 63.9\% & 51.5\% & 2.4\% & 4.3\% & 28,519 & 190\\
      Accuracy-Based Reward Policy & 64.4\% & 48.7\% & 4.2\% & 2.8\% & 52,829 & 348\\
      \hline
    \end{tabular}
    \caption{This table shows the CNN test and finetuned model accuracies for different output image sets. The Pure Chance Baseline shows the odds of pure chance (with the public task having 2 classes and the private task having 36). The Quantum Augmentation Baseline are the CNN models used for finetuning, and shows the test accuracies of the models trained on EMNIST data run through random quantum privacy gates. The Quantum Random Action Baseline was generated by randomly selecting actions from the quantum action space and simulating the output. The Gaussian Blur Test Baseline consists of a dataset with a 4x4 kernel Gaussian blur applied to each image. The Gaussian Noise Test Baseline consists of a dataset with normalized Gaussian noise added to each image. The rows following the baselines represent tests run on the image outputs of selected RL policies. The best performing model according to this metric, the Public-Based Reward Policy, is bolded for clarity.}
    \label{fig:results-table}
\end{figure*}

\section{Agent Learning}

Reinforcement learning was selected in order to fully explore and develop a good policy for selecting privacy preserving quantum actions. A decision on what action to take needs to be made for each image that comes through the simulated quantum camera. 

A reward can be constructed to promote actions resulting in public utility and to punish actions resulting in access to sensitive information. The agent can explore the space and fully take advantage of the actions available to maximize the reward and achieve the desired privacy-utility ratio.

\textbf{Reinforcement Learning with DDQN} Double Deep Q Learning (DDQN) is a type of reinforcement Q-learning that results in less overestimation and better performance than the standard Double and Deep Q-Learning (DQN) algorithms\cite{hasselt_deep_2016}. Reinforcement learning’s goal is to learn good policies so an agent can successfully navigate in its environment. It is popular for video games and robotics, where the agent is the game character or robot. The environment in this case is the “world” the agent navigates, and the action space is composed of the things the agent can do like move and jump. For imaging tasks, the environment can be thought of as the imaging system and picture target while the action space can be thought of as the controls the agent can access to manipulate the camera. 

To use DDQN in a quantum privacy preserving camera setup, the environment is created to include the "quantum camera" (in our case, represented by an FRQI encoder), privacy gate controls, and CNNs responsible for grading the measured image. The agent includes the target and online networks. The action space covers all combinations of the quantum privacy gates allowed in a single step. With this infrastructure, the DDQN model can learn a policy that maximizes the public utility task (in our case, the accuracy of the public classification task) while preserving privacy by minimizing the private task (the accuracy of the private classification task).

We demonstrate that DDQN can learn policies for a quantum camera framework by showing model metrics. Fig. \ref{fig:graph-table} shows the target and online Q values converge and loss decrease, indicating a policy is being learned. The stability of the reward indicates that the model is not overoptimistic.

\section{Experimental Design}

When designing our privacy preserving quantum framework, we assume temporal consistency. This assumption means thinking of the images like frames in a video, where the letters and numbers contain similar features across time steps. An example of this could be the camera scanning documents containing text. The agent sequentially chooses an action to apply to each character in the document, aiming to preserve the public utility features while redacting the sensitive private features. 

\textbf{Training Methodology} The DDQN architecture was used to realize a privacy preserving quantum camera. The agent produces an action that corresponds to which quantum gates are applied to an unknown FRQI image. Once the action gates are applied, the FRQI image is measured and decoded. The resulting damaged image is sent to two CNNs trained on augmented EMNIST data. The data augmentation included downscaling the images for speed and training on common image results from the action gates. One CNN is for identifying the public classes of “letter” and “number,” and one CNN is for identifying the private classes corresponding to each specific letter and number. The reward is determined by CNN accuracy or episode length, depending on the policy. The last two layers of each CNN and the reward are used by the DDQN networks to determine the current and target Q values, as well as the next action to be taken.

\textbf{Testing Methodology} After training a policy, the DDQN agent networks were frozen to assess their performance. This was done by running a test set through the model and saving the generated output image along with the ground truth label at each step. Once the images were generated, the quality of the policy was assessed using the privacy and public utility CNNs to obtain overall testing accuracies. These test accuracies were recorded, then the private and public utility CNNs were finetuned on the generated data, retested, and recorded.

This testing methodology differs slightly from testing methods presented in the original DDQN paper, which relies mainly on reward scores and comparing the learned policy to an empirical best policy\cite{vanhasseltDeepReinforcementLearning2015}. For this use case, we must show that the learned policy is empirically better than chance for it to be considered good. Since we construct and test several different reward functions (as opposed to using a predefined score like in a video game), we must also show that a higher reward corresponds with a better policy outcome. We chose the private and public utility CNN accuracies as a clear, empirical way of comparing the quality of a learned policies and the outcomes of different reward functions. 

\subsection{Results}

Different RL parameters and rewards were experimented with. Four models were chosen to be tested and the classification CNNs were subsequently finetuned on the image data produced by the tests. The competing classifier CNN models were run on NVIDIA GeForce GTX 1080 Ti's while the DDQN models were run on NVIDIA TITAN RTX's or NVIDIA TITAN Xp's. The results of these tests can be found in Fig. \ref{fig:results-table}. 

The Public-Based Reward Policy gives a set reward whenever the CNN can accurately predict the public utility label. This model achieved the highest performance, as the desired public utility accuracy was above chance while the private task accuracy was below chance. It was most similar to the Gaussian Noise Augmented Baseline for the public utility task, but achieved significantly better performance compared to the baseline for the private task. Additionally, the Gaussian Noise Augmented Baseline was able to be improved with finetuning while the Public-Based Reward Policy was not improved with finetuning.

The Public-Private Reward Policy add points to the reward when the CNNs can accurately predict the public utility label, but also subtracts points from the reward when the CNNs can accurately predict the private label. The Length-Based Reward Policy awards points based on how long the episode has been running above an accuracy threshold, thereby promoting longer episode times. The Accuracy-Based Reward Policy awards points based on the current episode public utility accuracy, thereby promoting higher public accuracies. These models were able to keep the private task accuracy at or near chance. However, the finetuned public utility accuracy was lower than the accuracy of the Public-Based Reward Policy.

\section{Discussion and Limitations}
The best performing model was the Public-Based Reward Policy, as it succeeded in keeping the images private and the public task accuracy better than random. This is significant given the state-of-the-art. In \cite{pittalugaLearningPrivacyPreserving2019}, which is a conventional, purely silicon-based deep learning method, the difference between their public baseline (blur, 86.9\%) and their proposed method (learned encoder, 91.6\%) was 4.7\%. Privacy is a hard problem, and we have taken the first step into quantum-based privacy preservation. In light of this, on-par performance with the public task (while doing better on the private task) is a reasonable performance.

We are the first to demonstrate that quantum gates can be used, in concert with a machine learning algorithm, to learn privacy preserving quantum actions for images. However, the sub-field we are investigating here has many challenges that we document, and there are two main limitations:

\begin{enumerate}
    \item Quantum cameras are currently in the early stages of development, and real quantum images are extremely low-resolution. Therefore simulations have resulted in long training and test times. 
    \item While our method shows promise, currently we are only slightly better than augmenting images with Gaussian noise. However, due to the sophisticated nature of quantum actions, we believe that, in the future, we will be able to fully exploit the action space to increase privacy protection. 
\end{enumerate}

We have selected quantum circuits for privacy, integrated QImP terminology into a machine learning framework and written code for our novel design that we will release so that other researchers may build on our initiative. Quantum computing in vision will have many impacts, and this paper shows that privacy preservation will be one of them. 

{\small
\bibliographystyle{ieee_fullname}
\bibliography{egbib,references,sanjeev}
}

\end{document}